%% file: paper.tex
\documentclass{article}

\input{preamble}

\usepackage{xcolor}

\title{Evaluating Large Language Models on Solved and Unsolved Problems in Graph Theory: Implications for Computing Education}

\author{
Adithya Kulkarni$^{\dagger}$,
Mohna Chakraborty$^{\ddagger}$,
and Jay Bagga$^{\dagger}$\\
$^{\dagger}$Department of Computer Science\\
Ball State University\\
Muncie, IN 47306\\
\email{\{adithya.kulkarni,jbagga\}@bsu.edu}\\
$^{\ddagger}$School of Artificial Intelligence and Data Science\\
Jio Institute\\
Navi Mumbai, India\\
\email{mohna.chakraborty@jioinstitute.edu.in}
}
\begin{document}
\maketitle

\begin{abstract}
Large Language Models are increasingly used by students to explore advanced material in computer science, including graph theory. As these tools become integrated into undergraduate and graduate coursework, it is important to understand how reliably they support mathematically rigorous thinking. This study examines the performance of a LLM on two related graph theoretic problems: a solved problem concerning the gracefulness of line graphs and an open problem for which no solution is currently known. We use an eight stage evaluation protocol\footnote{The prompts and LLM outputs can be found at \url{https://github.com/kulkarniadithya/Dual_Role_of_LLM/}} that reflects authentic mathematical inquiry, including interpretation, exploration, strategy formation, and proof construction.

The model performed strongly on the solved problem, producing correct definitions, identifying relevant structures, recalling appropriate results without hallucination, and constructing a valid proof confirmed by a graph theory expert. For the open problem, the model generated coherent interpretations and plausible exploratory strategies but did not advance toward a solution. It did not fabricate results and instead acknowledged uncertainty, which is consistent with the explicit prompting instructions that directed the model to avoid inventing theorems or unsupported claims.

These findings indicate that LLMs can support exploration of established material but remain limited in tasks requiring novel mathematical insight or critical structural reasoning. For computing education, this distinction highlights the importance of guiding students to use LLMs for conceptual exploration while relying on independent verification and rigorous argumentation for formal problem solving.
\end{abstract}

\section{Introduction}

Large Language Models (LLMs) are increasingly used by students in computer science courses to clarify concepts, explore ideas, and assist with problem solving. Prior work documents both the potential educational benefits of these tools and the risks associated with overreliance on model generated reasoning \cite{wang2024large,sharma2025role,guizani2025systematic,yan2024practical}. In courses involving mathematically rigorous material, such as graph theory, understanding how LLMs reason is essential for responsible integration into computing education.

Graph theory provides a useful setting for examining these issues. It is a foundational subject in many computing curricula and supports applications in networks, data science, and machine learning. Within graph theory, line graphs and graceful labeling are well studied topics that remain challenging even at advanced levels \cite{bagga2004old,bagga2021survey,gallian2018dynamic}. Students increasingly consult LLMs when engaging with such material, which motivates a careful evaluation of model behavior in this domain.

This study focuses on two closely related problems in graph theory that differ in logical direction and in their status within the literature. The first problem, which is solved, asks whether the line graph of a nongraceful graph must be graceful:
\[
\textbf{Problem 1:} \quad \text{If $G$ is nongraceful, is $L(G)$ graceful?}
\]
The second problem, which remains open, asks whether the converse implication holds:
\[
\textbf{Problem 2:} \quad \text{If $L(G)$ is graceful, is $G$ graceful?}
\]
While these statements appear superficially similar, they pose fundamentally different challenges. Problem 1 has known counterexamples and established solutions in the literature, allowing us to evaluate whether a model can correctly interpret definitions and reproduce valid reasoning. In contrast, no general solution is known for Problem 2, even for restricted classes of graphs, making it a suitable test of model behavior in the presence of genuine mathematical uncertainty.

We evaluate the model using an eight stage protocol that reflects the typical phases of mathematical inquiry. The model is asked to interpret each problem, generate exploratory ideas, identify related areas, recall relevant results, propose strategies, construct proofs, critique its own reasoning, and revise its arguments. This structure enables a detailed analysis of both exploratory reasoning and formal deduction.

The contributions of this study are threefold. First, we provide an empirical characterization of LLM behavior on both a solved and an unsolved problem in graph theory. Second, we show that the model performs reliably on tasks grounded in established theory, producing accurate interpretations and a correct proof for the solved problem that was confirmed by an expert. Third, we show that for the open problem, the model generates coherent exploratory reasoning but does not produce new mathematical insight or advance the state of knowledge.

Overall, this work clarifies how LLMs can support conceptual exploration in advanced computing courses while also highlighting their limitations in tasks that require verification, abstraction, and original reasoning.

\section{Background and Related Work}

This section reviews prior work relevant to three aspects of this study: graph theory and line graphs, the use of LLMs in educational contexts, and research on hallucination and reasoning behavior in LLMs. Together, these areas provide the context for evaluating how an LLM behaves when engaging with both solved and unsolved problems in graph theory.

\subsection{Graph Theory, Line Graphs, and Graceful Labeling}

Line graph theory is a well established area of graph theory with foundational results dating to the mid twentieth century and continued development in recent decades. Research in this area documents both classical properties and modern generalizations of line graphs, reflecting their broad applicability and sustained theoretical interest~\cite{bagga2004old,bagga2021survey}. Graceful labeling is another central topic in graph theory, characterized by a large and evolving body of literature. Gallian provides a comprehensive and continuously updated survey of graph labeling results, including graceful graphs and related constructions~\cite{gallian2018dynamic}. Both line graphs and graceful labeling are frequently encountered in advanced graph theory courses and remain active areas of research.

Additional work has examined structural properties and construction techniques for graceful graphs~\cite{cavalier2009graceful,johnson2018constructing}. More recent studies explore line graph transformations in computational settings such as network analysis and machine learning, demonstrating the continued relevance of these concepts beyond purely theoretical contexts~\cite{yang2024theoretical,ming2025deep}. The depth and diversity of this literature make line graphs and graceful labeling well suited for evaluating reasoning in mathematically rigorous settings.

\subsection{Large Language Models in Education and Research}

LLMs are increasingly used in higher education to support conceptual understanding, literature exploration, assignment assistance, and programming tasks. Surveys and empirical studies report potential benefits such as improved accessibility, personalized learning support, and increased efficiency in instructional contexts~\cite{wang2024large,sharma2025role,guizani2025systematic,lopez2025integrating}. Additional research examines the broader lifecycle of LLM adoption in education, including issues related to bias, transparency, and long term impact on learning practices~\cite{lee2024life}.

At the same time, concerns have been raised about students’ tendency to overestimate the reliability of LLM generated output, particularly for tasks involving complex reasoning. Studies in STEM education document cases in which learners accepted incorrect or incomplete explanations produced by LLMs, leading to misunderstandings or the reinforcement of errors~\cite{krupp2023unreflected,li2024impact}. These findings motivate closer examination of how LLMs behave in mathematically demanding contexts, where correctness depends on precise definitions and rigorous logical structure.

\subsection{Hallucination and Reasoning Behavior in LLMs}

A substantial body of work investigates hallucination and reasoning limitations in LLMs. Recent surveys categorize different forms of hallucination, analyze contributing factors, and discuss the implications for reliable deployment of these systems~\cite{huang2025survey,alansari2025large}. In mathematical and logical domains, LLMs have been shown to generate incorrect statements, misapply known results, or construct arguments that violate formal reasoning principles~\cite{boye2025large,ahn2024large,xu2024hallucination,heyman2025reasoning}. Such behavior raises concerns about the suitability of LLMs for tasks requiring deductive validity.

Other studies examine internal signals associated with hallucination risk~\cite{ji2024llm,orgad2024llms} and analyze the educational consequences of erroneous AI generated feedback~\cite{steinbach2025llms}. Proposed mitigation approaches include self consistency checks, structured reasoning prompts, and fine grained hallucination detection techniques~\cite{liu2025enhancing,mukherjee2025premise,li2024fg,li2025mitigating,ouyang2025treecut}. While these methods improve reliability in some settings, they do not eliminate the fundamental limitations of current models when reasoning requires original insight or extrapolation beyond known patterns.

Taken together, prior work suggests that LLMs are effective at organizing and explaining established material but remain limited when tasks require the construction of new mathematical arguments. This distinction motivates our evaluation of LLM behavior on both a solved and an unsolved problem in graph theory and informs the educational implications discussed in the remainder of this paper.

\section{Methodology}

For a graph $G$, the line graph $L(G)$ has a vertex for each edge of $G$, and two vertices in $L(G)$ are adjacent if the corresponding edges of $G$ are adjacent \cite{beineke2021line}. This study examines the behavior of a LLMs when engaged in a structured mathematical problem solving workflow involving two closely related statements about gracefulness and the line graph operator.  The two statements differ in logical direction and in their status in the literature. Problem 1 is solved, meaning that established arguments and known counterexamples exist. Problem 2 is open, meaning that no general proof or counterexample is currently known. This design supports a direct comparison between model behavior when a canonical resolution exists and when genuine mathematical uncertainty remains.

\subsection{Why Problem 1 Is Solved and Why Problem 2 Is Difficult}

Problem 1 asks:
\[
\textbf{Problem 1:} \quad \text{If $G$ is nongraceful, is $L(G)$ graceful?}
\]
This statement is solved because the implication is false: there are graphs $G$ that are nongraceful for which $L(G)$ is not graceful. In other words, counterexamples exist, so the statement can be resolved by exhibiting a specific $G$ and verifying that neither $G$ nor $L(G)$ admits a graceful labeling. This type of resolution is typical in graph labeling, where disproving an implication often reduces to identifying a concrete graph family with a provable obstruction.

Problem 2 asks the converse implication:
\[
\textbf{Problem 2:} \quad \text{If $L(G)$ is graceful, is $G$ graceful?}
\]
Problem 2 remains difficult because it requires reasoning about how graceful labelings behave under an inverse transformation. The line graph operator loses information: different nonisomorphic graphs can share the same line graph structure, and structural properties of $L(G)$ do not uniquely determine labeling properties of $G$. As a result, techniques that work for forward implications often do not transfer to the reverse direction. In practice, progress on Problem 2 typically requires either a general structural theorem that lifts labelings from $L(G)$ back to $G$ or a construction of a counterexample family. Neither is currently known.

\subsection{Experimental Environment and Materials}

All experiments were conducted using the publicly available web interface of ChatGPT 5.1\footnote{\url{https://openai.com/index/gpt-5-1/}.}. The interface supports multimodal inputs such as PDF uploads, which is consistent with common student practice in advanced coursework. Students frequently provide a problem statement together with lecture notes or papers and then ask the model to interpret definitions, suggest research directions, or critique partial solutions.

In the first prompt for each problem, the model received two background documents: \textit{A Dynamic Survey of Graph Labeling} \cite{gallian2018dynamic} and \textit{On Graceful Line Graphs}\footnote{This document was obtained from the authors at the 2025 CCSC Midwest Conference and was used as background material for definitions and context.}. These documents provide definitions and known results related to line graphs and graceful labelings. Each problem was evaluated in a separate conversation to avoid cross conversation persistence. All prompts and model outputs were saved verbatim. No feedback, correction, or reinforcement was provided during any stage of the evaluation.

\subsection{Evaluation Protocol and Prompt Design}

The evaluation uses an eight stage protocol that reflects a typical workflow students follow when using LLMs for difficult mathematical tasks: first clarifying the problem and definitions, then generating ideas and relevant context, and only later attempting formal proof and revision. The same prompt templates were used for both problems, with only the underlying problem statement changed. Several prompts explicitly instructed the model to avoid inventing theorems or unsupported claims and to indicate uncertainty when appropriate, in order to reduce fabricated academic content and to observe model behavior under this constraint.

For completeness, the exact wording of the prompts as presented to the model for Problem 1 is provided below. Italics indicate the text displayed to the model.

\medskip
\textbf{Prompt 1: Problem Understanding.}  

\textit{I want your help in developing proof (either prove or disprove) for the below unsolved problem:}

\textit{Problem: If $G$ is nongraceful then $L(G)$ is graceful?}

\textit{I am providing two papers that have information on the definitions of line graph, graceful labeling, and related concepts. These papers also provide you with necessary background that can be useful to develop the proof for the problem.}

\textit{Please understand the problem and read the attached PDFs carefully and thoroughly. Then:}
\begin{enumerate}
    \item \textit{Restate the main problem or conjecture in your own words.}
    \item \textit{List all key definitions and properties involved.}
    \item \textit{Explain why the problem is challenging or interesting.}
    \item \textit{Identify any natural simplifications or special cases that might be useful starting points.}
\end{enumerate}

\medskip
\textbf{Prompt 2: Brainstorming.}  

\textit{Based on your understanding of the problem, please brainstorm 6--10 possible approaches or ideas that a researcher might explore. For each idea:}
\begin{enumerate}
    \item \textit{Describe the core insight.}
    \item \textit{Explain why it might be promising.}
    \item \textit{Point out potential obstacles.}
\end{enumerate}

\medskip
\textbf{Prompt 3: Related Areas.}  

\textit{Before we proceed, can you identify the areas of graph theory or graph labeling that are most closely related to the problem. For each area:}
\begin{enumerate}
    \item \textit{Name it.}
    \item \textit{Explain the relationship to the problem.}
    \item \textit{Mention concepts or tools from that area that might be relevant.}
\end{enumerate}

\medskip
\textbf{Prompt 4: Theorem and Result Recall.}  

\textit{Now list 8--12 theorems, known results, or named conjectures that you believe are directly relevant to the problem. For each one:}
\begin{enumerate}
    \item \textit{Provide the name of the result.}
    \item \textit{State the authors (if known).}
    \item \textit{Give a 2--3 sentence summary.}
    \item \textit{Indicate if you are unsure.}
\end{enumerate}
\textit{Please do not invent results or authors. Only include items you have high confidence in.}

\medskip
\textbf{Prompt 5: Strategy Formation.}  

\textit{Propose 3--5 distinct high level strategies to try to resolve the problem. They do not need to be correct. For each strategy:}
\begin{enumerate}
    \item \textit{Explain the core idea.}
    \item \textit{List any structural lemmas that might be needed.}
    \item \textit{Predict the hardest obstacle in making the strategy succeed.}
\end{enumerate}

\textbf{Prompt 6: Proof Attempt.}  

\textit{Now attempt to prove the main conjecture or implication. Write the proof in a rigorous, step by step manner, as if submitting to a mathematics journal. State each lemma clearly, justify each inference, and avoid vague reasoning.}

\medskip
\textbf{Prompt 7: Self Evaluation.}  

\textit{Please re read your entire proof carefully. Identify any steps that may be incorrect, incomplete, or unjustified. For each such step:}
\begin{enumerate}
    \item \textit{Explain why it might be flawed.}
    \item \textit{Attempt to repair or clarify the argument.}
\end{enumerate}

\textbf{Prompt 8: Revised Proof.}  

\textit{Now, can you correct the identified flaws and provide a final rigorous proof for the problem without any flaws.}

\subsection{Expert Evaluation Procedure}

All outputs were evaluated by a faculty expert in graph theory with research experience in line graphs and graph labelings. The expert was not an expert in artificial intelligence, which reduces the risk that judgments were influenced by assumptions about typical LLM behavior. Prior to evaluation, the expert was provided with explicit written instructions specifying what aspects of each prompt response to analyze. These prompt specific evaluation guidelines were designed to align with the goals of each stage of the protocol and are publicly available in the project repository.

For each prompt and each problem, the expert recorded qualitative comments and assessed the LLM output against the provided criteria. For prompts focused on interpretation and exploration (Prompts 1 to 5), the expert evaluated conceptual correctness of definitions, fidelity of problem restatement, relevance of identified themes, and appropriateness of proposed research directions. For the theorem recall prompt (Prompt 4), the expert verified whether listed theorems and author attributions were correct, relevant, or potentially fabricated. For proof oriented prompts (Prompts 6 to 8), the expert evaluated deductive validity by checking whether claims followed logically from prior statements, whether definitions were applied correctly, and whether the argument addressed the stated problem. The expert also examined whether the model’s self evaluation and revision stages identified substantive flaws and made meaningful corrections.

This evaluation procedure supports a systematic comparison between the model's exploratory usefulness and its limitations in proof oriented reasoning. It also enables a direct contrast between performance on a solved problem, where established reasoning exists, and an unsolved problem, where no reference solution is available.

\section{Results}

The evaluation examined the model's behavior on two graph theoretic problems that differ in logical direction and in their status within the literature: a solved problem and an unsolved problem. Expert analysis shows that the model performed reliably when established solutions and reasoning patterns were available, while its limitations became apparent when the task required reasoning beyond known mathematical results. The findings are presented by problem and by stage of the eight prompt evaluation protocol.

\subsection{Problem 1: If $G$ is nongraceful, is $L(G)$ graceful?}

\subsubsection{Prompts 1 to 5: Exploratory and Conceptual Reasoning}

Across the first five prompts, the model demonstrated strong conceptual understanding and generated reasoning that aligned closely with expert expectations.

\noindent \textbf{Prompt 1: Problem Interpretation.}  
The model restated the problem accurately, identified all relevant definitions, and extracted the structural role of the line graph transformation. The expert noted that the interpretation was precise and reflected a correct understanding of the conjecture.

\noindent \textbf{Prompt 2: Idea Generation.}  
The model produced several correct and relevant approaches, including use of degree patterns, properties of known graceful families, and the examination of structural counterexamples. The expert found these ideas appropriate and aligned with genuine mathematical strategies used in the literature.

\noindent \textbf{Prompt 3: Related Areas.}  
The model identified areas such as graph labelings, structural characterizations of line graphs, tree labelings, and decomposition methods. The expert evaluation confirmed the relevance of these areas and the correctness of the connections drawn.

\noindent \textbf{Prompt 4: Recall of Results.}  
A notable outcome is that the model did not hallucinate any theorems. All referenced results were correct, relevant, and stated with appropriate caution. This behavior contrasts with patterns documented in prior work and demonstrates unusually reliable performance on this prompt.

\noindent \textbf{Prompt 5: Strategy Formation.}  
The strategies were coherent, structurally sound, and consistent with how a human expert might begin approaching the problem. The expert noted that the model displayed a clear understanding of how to decompose the problem into meaningful subproblems.

\subsubsection{Prompts 6 to 8: Proof Construction and Self Evaluation}

\noindent \textbf{Prompt 6: Proof Attempt.}  
The model produced a correct argument for the solved problem. It identified a counterexample and provided a complete proof that the expert confirmed as valid. No circular reasoning, fabricated lemmas, or unjustified generalizations were observed.

\noindent \textbf{Prompt 7: Self Evaluation.}  
The model reviewed its proof and did not introduce new errors. The expert indicated that the self evaluation was reasonable given that the proof was already correct.

\noindent \textbf{Prompt 8: Revised Proof.}  
The model presented a refined version of the proof that remained correct. The expert confirmed that the proof was complete, logically sound, and consistent with known results.

Overall, for Problem 1 the model demonstrated reliable conceptual reasoning, accurate recall of known results, and correct formal reasoning.

\subsection{Problem 2: If $L(G)$ is graceful, is $G$ graceful?}

\subsubsection{Prompts 1 to 5: Exploratory and Conceptual Reasoning}

The model again showed strong performance on conceptual and exploratory tasks, despite the fact that the problem is open and has no known resolution.

\noindent \textbf{Prompt 1: Problem Interpretation.}  
The model restated the problem clearly and identified correct definitions and structural components. It also articulated the reasons the conjecture is difficult, such as limited understanding of inverse line graph behavior in the context of graceful labelings. The expert noted that the explanation was accurate and appropriate.

\noindent \textbf{Prompt 2: Idea Generation.}  
The model proposed several useful exploratory ideas, including analysis of special graph families and examination of structural features in the inverse direction of the line graph operation. While some ideas were incomplete or required refinement, the expert found them reasonable and aligned with authentic research directions.

\noindent \textbf{Prompt 3: Related Areas.}  
The model correctly connected the problem to labeling theory, line graph characterizations, and decomposition methods. The expert evaluation confirmed that these connections were sound.

\noindent \textbf{Prompt 4: Recall of Results.}  
As in Problem 1, the model did not hallucinate any theorems. The results it listed were correct and relevant. This consistency across two problems suggests reliability in recalling established literature.

\noindent \textbf{Prompt 5: Strategy Formation.}  
The model produced coherent strategies for investigating the conjecture, including systematic examination of special graph classes and potential structural reductions. The expert found these strategies meaningful and aligned with the type of exploratory work expected in early research stages.

\subsubsection{Prompts 6 to 8: Proof Construction and Self Evaluation}

\noindent \textbf{Prompt 6: Proof Attempt.}  
The model attempted to construct a proof but did not succeed. Importantly, its argument did not contain fabricated theorems or logically incoherent steps. Instead, the expert observed that the model produced partial reasoning that remained bounded by known results and did not invent unsupported claims.

\noindent \textbf{Prompt 7: Self Evaluation.}  
The model correctly identified its own limitations, noting instances of incomplete reasoning and uncertainty. The expert described this as a reasonable and honest assessment of its own output.

\noindent \textbf{Prompt 8: Revised Proof.}  
The model did not claim to have resolved the open problem. Instead, it shifted toward describing partial progress and suggesting relevant subproblems. The expert found this appropriate and remarked that the model accurately refrained from fabricating a solution.

For Problem 2, the model demonstrated strong exploratory skills but was unable to produce new mathematical insight, which is expected given the open nature of the problem.

\noindent \textbf{Summary}  
Across both problems, the model showed reliable conceptual and exploratory reasoning. It performed particularly well on the solved problem, producing a correct proof and accurate references, which demonstrates an ability to organize and apply established knowledge. For the open problem, the model again displayed coherent exploratory reasoning and did not fabricate results or claim a solution, instead acknowledging uncertainty. At the same time, its performance revealed a consistent limitation: the model was unable to engage in the type of critical thinking required to generate new structural insight or evaluate the deeper consequences of its own partial arguments. This inability to move beyond known patterns and to reason flexibly in unfamiliar settings highlights a fundamental constraint of current Large Language Models. These findings suggest that while the model is effective for engaging with established material, it remains limited in tasks that require original mathematical discovery, critical evaluation, or synthesis beyond existing knowledge.

\section{Discussion}

The results reveal a clear contrast between the model's behavior on a solved problem and an unsolved one. When established theory was available, the model interpreted the problem accurately, recalled relevant results, and constructed a valid proof. In contrast, for the unsolved problem, the model produced coherent exploratory reasoning but did not generate new insight or extend existing results. This distinction has important implications for computing education as students increasingly use LLMs in mathematically rigorous settings.

\subsection{Implications for Student Use of LLMs}

The model’s performance on the exploratory stages of both problems indicates that these systems can be useful in helping students begin work on complex or unfamiliar material. The model was able to restate each conjecture accurately, identify relevant definitions and structures, and propose plausible strategies. These capabilities are consistent with prior research showing that LLMs can support conceptual understanding and reduce initial barriers in advanced coursework \cite{wang2024large,sharma2025role,guizani2025systematic}.

The inability of the model to advance the unsolved problem, however, illustrates a limitation that students may overlook. Although the model did not hallucinate theorems or assert false conclusions, it also did not perform the type of reasoning that produces new mathematical ideas. Students may interpret a coherent explanation as evidence of underlying understanding, even when the model is only reorganizing existing patterns. Prior work documents that learners may accept fluent but unverified reasoning as authoritative \cite{krupp2023unreflected,steinbach2025llms}. Students therefore need explicit guidance to recognize the distinction between exploratory assistance and reliable mathematical reasoning.

\subsection{Critical Thinking Limitations of LLMs}

The distinction between the solved and unsolved problems provides a focused setting in which to observe the limitations of current models in tasks that require critical thinking. The model did not propose new constructions, search for counterexamples, refine partial ideas based on structural consequences, or assess the implications of its own reasoning steps. These behaviors are central to mathematical discovery but are not part of the inference process used by LLMs.

LLMs generate responses by estimating patterns in training data rather than by performing deliberate analysis or hypothesis testing. They do not build abstractions, evaluate alternative possibilities, or engage in reflective reasoning about correctness. As a result, they can reproduce reasoning that resembles known arguments but cannot reliably extend those arguments into new domains. This structural limitation explains why the model performed well on the solved problem but could not progress on the unsolved one, despite producing coherent and relevant intermediate ideas.

\subsection{Implications for Instruction and Course Design}

For instructors, the results suggest that LLMs can be integrated productively into the early phases of learning. Tasks involving conceptual restatement, identification of related areas, or preliminary strategy formation align with the strengths observed in the evaluation. Activities that require students to compare their own ideas with model generated ideas or to critique these suggestions may support metacognitive development.

It is equally important to design assignments that require students to verify claims, assess the structure of arguments, and recognize the limitations of model generated reasoning. The inability of the model to produce novel insight on the unsolved problem demonstrates that these tools cannot replace the critical analysis that is central to advanced study in graph theory and related areas. Instruction should therefore emphasize the need to evaluate model output independently and to understand the boundary between known results and open problems.

\subsection{Relevance for Advanced Topics in Computing}

Graph theory remains an essential area of instruction in many computer science programs, with continued research activity and broad applicability \cite{bagga2004old,bagga2021survey}. As students encounter topics that approach current research frontiers, they may increasingly rely on LLMs to clarify definitions or explore structural themes. The findings of this study indicate that such use can be productive when the material is grounded in established knowledge, but that these tools cannot be expected to resolve open questions or generate new theoretical contributions.

\medskip
\noindent \textbf{Summary}
Overall, the findings show that LLMs are effective exploratory tools but remain limited as reasoning engines in tasks that require original mathematical insight. Their strengths can support conceptual engagement, problem interpretation, and strategic planning, but they must be complemented by explicit training in verification and mathematical rigor. Responsible integration in computing education requires helping students understand both the capabilities and the boundaries of these systems.

\section{Recommendations}

The results of this study show that LLMs are effective for exploratory engagement with established material but remain limited when tasks require verification or novel mathematical reasoning. Based on these findings, we offer concrete recommendations for students, instructors, and researchers who use LLMs in advanced computing contexts.

\subsection{For Students}

\noindent \textbf{Use LLMs to initiate problem solving, not to conclude it.}  
Students can use LLMs to restate problems, recall definitions, identify relevant concepts, and generate possible approaches. However, correctness must be established through independent reasoning, as demonstrated by the model’s inability to resolve the unsolved problem.

\noindent \textbf{Treat model output as a draft, not an answer.}  
When reviewing LLM generated arguments, students should explicitly check each logical step, verify assumptions, and confirm whether conclusions follow from definitions.

\noindent \textbf{Use LLMs to orient literature exploration.}  
LLMs can help students identify relevant terminology, subtopics, and canonical references that serve as starting points for literature searches. This use is most effective when model output is treated as an initial guide and all references are verified using authoritative academic sources.

\noindent \textbf{Verify claims using external sources.}  
References to theorems or known results should be checked against textbooks, peer reviewed literature, or instructor provided materials. Prior studies show that students may accept plausible but incorrect reasoning unless verification is explicitly required~\cite{krupp2023unreflected,steinbach2025llms}.

\subsection{For Instructors}

\noindent \textbf{Integrate LLMs into early phases of coursework.}  
The model performed consistently well in tasks such as problem interpretation and idea generation. Instructors can incorporate LLMs into activities that emphasize understanding problem structure or exploring alternative approaches.

\noindent \textbf{Require explicit validation of LLM generated reasoning.}  
Assignments should ask students to assess the correctness of model output, identify limitations, or explain why an argument succeeds or fails.

\noindent \textbf{Set clear expectations for appropriate use.}  
Explicit guidelines can help students recognize when LLMs are useful and when independent analysis is required, particularly for open problems or proof based tasks.

\subsection{For Researchers}

\noindent \textbf{Use LLMs for exploration and organization.}  
The model generated coherent strategies and contextual information but did not produce new insight for the open problem. Researchers should treat LLMs as tools for organizing known material rather than resolving open questions.

\noindent \textbf{Use LLMs as orientation tools for unfamiliar areas.}  
LLMs may assist researchers in identifying common terminology, related problem areas, and possible connections across subfields. This role is limited to initial orientation and should precede systematic literature review.

\noindent \textbf{Maintain transparency and verification.}  
When LLMs are used during exploratory stages, prompts and outputs should be documented and all claims independently verified.

Overall, these recommendations emphasize responsible and practical use of LLMs in mathematically rigorous settings. LLMs can support orientation and exploration, but verification and critical reasoning must remain central to learning and research.

\section{Conclusion}

This study evaluated how an LLM responded to two graph theoretic problems, one solved and one unsolved, using an eight stage protocol that examined conceptual reasoning, strategy formation, and proof construction. The results show that the model performs reliably when the problem has an established solution. For the solved problem, it interpreted the conjecture accurately, recalled relevant results, and produced a correct proof confirmed by an expert. These findings indicate that LLMs can assist learners in understanding definitions, recognizing structures, and engaging with known material.

For the unsolved problem, the model offered coherent interpretations and exploratory ideas but did not generate new insight or progress toward a solution. It did not fabricate results and instead acknowledged uncertainty. This behavior reflects a structural limitation of current models: they can organize and apply existing knowledge but cannot perform the critical thinking or abstraction required for novel mathematical reasoning.

These observations have direct implications for computing education. LLMs can support early stage exploration but cannot replace verification or formal argumentation. Effective use in advanced coursework requires guiding students in validating model output, understanding its limits, and developing independent reasoning skills. Overall, the findings highlight both the value and the boundaries of LLMs as tools for learning in mathematically rigorous domains.

\printbibliography

\end{document}

%% file: preamble.tex
\usepackage[
  paperheight=8.5in,
  paperwidth=5.5in,
  left=10mm,
  right=10mm,
  top=20mm,
  bottom=20mm]{geometry}
\usepackage[utf8]{inputenc}

\usepackage{graphicx}
\usepackage{wrapfig}
\usepackage[bottom]{footmisc}
\usepackage{listings}
\usepackage{enumitem}

\usepackage{wrapfig}
\usepackage{ragged2e}

\usepackage{array}
\usepackage[table]{xcolor}
\usepackage{multirow}
\usepackage{booktabs}
\usepackage{hhline}
\definecolor{palegreen}{rgb}{0.6,0.98,0.6}

\usepackage{amsmath}
\usepackage{amssymb}
\usepackage{multicol}
\usepackage{lipsum}
\usepackage{hyphenat}
\PassOptionsToPackage{hyphens}{url}
\usepackage{url}

\usepackage{rotating}

\usepackage[T1]{fontenc}
\usepackage{textcomp}
\usepackage{listings}
\usepackage{setspace}

\newcommand*{\email}[1]{\small{\texttt{#1}}}

\renewcommand{\footnoterule}{%
  \kern -3pt
  \hrule width \textwidth height 0.5pt
  \kern 2pt
}

\date{}

\usepackage{titlesec}
\titleformat*{\section}{\large\bfseries}
\titleformat*{\subsection}{\normalsize\bfseries}
\titleformat*{\subsubsection}{\normalsize\bfseries}

\usepackage{biblatex}

%% file: references.bib
@book{beineke2021line,
  title={Line graphs and line digraphs},
  author={Beineke, Lowell and Bagga, Jay},
  year={2021},
  publisher={Springer}
}

@article{bagga2004old,
  title={Old and new generalizations of line graphs},
  author={Bagga, Jay},
  journal={International Journal of Mathematics and Mathematical Sciences},
  volume={2004},
  number={29},
  pages={1509--1521},
  year={2004},
  publisher={Wiley Online Library}
}

@article{bagga2021survey,
  title={A survey of line digraphs and generalizations},
  author={Bagga, Jay and Beineke, Lowell},
  journal={Discrete Math. Lett},
  volume={6},
  pages={68--83},
  year={2021}
}

@article{yang2024theoretical,
  title={Theoretical insights into line graph transformation on graph learning},
  author={Yang, Fan and Huang, Xingyue},
  journal={arXiv preprint arXiv:2410.16138},
  year={2024}
}

@article{ming2025deep,
  title={Deep Link Strength Prediction: Leveraging line graph transformations and neural networks},
  author={Ming, Zhixin and Li, Jie and Wang, Jing},
  journal={Journal of Computational Science},
  pages={102661},
  year={2025},
  publisher={Elsevier}
}

@article{gallian2018dynamic,
  title={A dynamic survey of graph labeling},
  author={Gallian, Joseph A},
  journal={Electronic Journal of combinatorics},
  volume={1},
  number={DynamicSurveys},
  pages={DS6},
  year={2018},
  publisher={Electronic Journal of Combinatorics}
}

@article{cavalier2009graceful,
  title={Graceful labelings},
  author={Cavalier, Charles Michael},
  journal={Bachelor of Science (Lousiana State University 2006) http://www. math. sc. edu/\~{} czabarka/Graceful. pdf},
  year={2009}
}

@article{johnson2018constructing,
  title={Constructing Graceful Graphs by Extending Paths from Graceful Graphs},
  author={Johnson, Kyrie},
  year={2018}
}

@article{huang2025survey,
  title={A survey on hallucination in large language models: Principles, taxonomy, challenges, and open questions},
  author={Huang, Lei and Yu, Weijiang and Ma, Weitao and Zhong, Weihong and Feng, Zhangyin and Wang, Haotian and Chen, Qianglong and Peng, Weihua and Feng, Xiaocheng and Qin, Bing and others},
  journal={ACM Transactions on Information Systems},
  volume={43},
  number={2},
  pages={1--55},
  year={2025},
  publisher={ACM New York, NY}
}

@article{liu2025enhancing,
  title={Enhancing mathematical reasoning in large language models with self-consistency-based hallucination detection},
  author={Liu, MingShan and Fang, Jialing},
  journal={arXiv preprint arXiv:2504.09440},
  year={2025}
}

@article{li2025mitigating,
  title={Mitigating Hallucination in Large Language Models (LLMs): An Application-Oriented Survey on RAG, Reasoning, and Agentic Systems},
  author={Li, Yihan and Fu, Xiyuan and Verma, Ghanshyam and Buitelaar, Paul and Liu, Mingming},
  journal={arXiv preprint arXiv:2510.24476},
  year={2025}
}

@article{wang2024large,
  title={Large language models for education: A survey and outlook},
  author={Wang, Shen and Xu, Tianlong and Li, Hang and Zhang, Chaoli and Liang, Joleen and Tang, Jiliang and Yu, Philip S and Wen, Qingsong},
  journal={arXiv preprint arXiv:2403.18105},
  year={2024}
}

@article{sharma2025role,
  title={The role of large language models in personalized learning: a systematic review of educational impact},
  author={Sharma, Sahil and Mittal, Puneet and Kumar, Mukesh and Bhardwaj, Vivek},
  journal={Discover Sustainability},
  volume={6},
  number={1},
  pages={1--24},
  year={2025},
  publisher={Springer}
}

@article{guizani2025systematic,
  title={A systematic literature review to implement large language model in higher education: issues and solutions},
  author={Guizani, Sghaier and Mazhar, Tehseen and Shahzad, Tariq and Ahmad, Wasim and Bibi, Afsha and Hamam, Habib},
  journal={Discover Education},
  volume={4},
  number={1},
  pages={1--25},
  year={2025},
  publisher={Springer}
}

@article{lopez2025integrating,
  title={Integrating Large Language Models into Accessible and Inclusive Education: Access Democratization and Individualized Learning Enhancement Supported by Generative Artificial Intelligence},
  author={Lopez-Gazpio, Inigo},
  journal={Information},
  volume={16},
  number={6},
  pages={473},
  year={2025},
  publisher={MDPI}
}

@article{lee2024life,
  title={The life cycle of large language models in education: A framework for understanding sources of bias},
  author={Lee, Jinsook and Hicke, Yann and Yu, Renzhe and Brooks, Christopher and Kizilcec, Ren{\'e} F},
  journal={British Journal of Educational Technology},
  volume={55},
  number={5},
  pages={1982--2002},
  year={2024},
  publisher={Wiley Online Library}
}

@article{alansari2025large,
  title={Large Language Models Hallucination: A Comprehensive Survey},
  author={Alansari, Aisha and Luqman, Hamzah},
  journal={arXiv preprint arXiv:2510.06265},
  year={2025}
}

@article{yan2024practical,
  title={Practical and ethical challenges of large language models in education: A systematic scoping review},
  author={Yan, Lixiang and Sha, Lele and Zhao, Linxuan and Li, Yuheng and Martinez-Maldonado, Roberto and Chen, Guanliang and Li, Xinyu and Jin, Yueqiao and Ga{\v{s}}evi{\'c}, Dragan},
  journal={British Journal of Educational Technology},
  volume={55},
  number={1},
  pages={90--112},
  year={2024},
  publisher={Wiley Online Library}
}

@article{ji2024llm,
  title={Llm internal states reveal hallucination risk faced with a query},
  author={Ji, Ziwei and Chen, Delong and Ishii, Etsuko and Cahyawijaya, Samuel and Bang, Yejin and Wilie, Bryan and Fung, Pascale},
  journal={arXiv preprint arXiv:2407.03282},
  year={2024}
}

@article{krupp2023unreflected,
  title={Unreflected acceptance--investigating the negative consequences of ChatGPT-assisted problem solving in physics education},
  author={Krupp, Lars and Steinert, Steffen and Kiefer-Emmanouilidis, Maximilian and Avila, Karina E and Lukowicz, Paul and Kuhn, Jochen and K{\"u}chemann, Stefan and Karolus, Jakob},
  journal={arXiv preprint arXiv:2309.03087},
  year={2023}
}

@inproceedings{li2024impact,
  title={The impact of large language models on higher education},
  author={Li, X and others},
  booktitle={Frontiers in Education},
  volume={9},
  year={2024}
}

@article{boye2025large,
  title={Large language models and mathematical reasoning failures},
  author={Boye, Johan and Moell, Birger},
  journal={arXiv preprint arXiv:2502.11574},
  year={2025}
}

@article{ahn2024large,
  title={Large language models for mathematical reasoning: Progresses and challenges},
  author={Ahn, Janice and Verma, Rishu and Lou, Renze and Liu, Di and Zhang, Rui and Yin, Wenpeng},
  journal={arXiv preprint arXiv:2402.00157},
  year={2024}
}

@article{xu2024hallucination,
  title={Hallucination is inevitable: An innate limitation of large language models},
  author={Xu, Ziwei and Jain, Sanjay and Kankanhalli, Mohan},
  journal={arXiv preprint arXiv:2401.11817},
  year={2024}
}

@article{heyman2025reasoning,
  title={Reasoning large language model errors arise from hallucinating critical problem features},
  author={Heyman, Alex and Zylberberg, Joel},
  journal={arXiv preprint arXiv:2505.12151},
  year={2025}
}

@article{mukherjee2025premise,
  title={Premise-augmented reasoning chains improve error identification in math reasoning with llms},
  author={Mukherjee, Sagnik and Chinta, Abhinav and Kim, Takyoung and Sharma, Tarun Anoop and Hakkani-T{\"u}r, Dilek},
  journal={arXiv preprint arXiv:2502.02362},
  year={2025}
}

@article{ouyang2025treecut,
  title={TreeCut: A Synthetic Unanswerable Math Word Problem Dataset for LLM Hallucination Evaluation},
  author={Ouyang, Jialin},
  journal={arXiv preprint arXiv:2502.13442},
  year={2025}
}

@article{li2024fg,
  title={FG-PRM: Fine-grained Hallucination Detection and Mitigation in Language Model Mathematical Reasoning},
  author={Li, Ruosen and Luo, Ziming and Du, Xinya},
  journal={arXiv preprint arXiv:2410.06304},
  year={2024}
}

@article{orgad2024llms,
  title={Llms know more than they show: On the intrinsic representation of llm hallucinations},
  author={Orgad, Hadas and Toker, Michael and Gekhman, Zorik and Reichart, Roi and Szpektor, Idan and Kotek, Hadas and Belinkov, Yonatan},
  journal={arXiv preprint arXiv:2410.02707},
  year={2024}
}

@inproceedings{steinbach2025llms,
  title={When LLMs Hallucinate: Examining the Effects of Erroneous Feedback in Math Tutoring Systems},
  author={Steinbach, Marlene and Bhandari, Shreya and Meyer, Jennifer and Pardos, Zachary A},
  booktitle={Proceedings of the Twelfth ACM Conference on Learning@ Scale},
  pages={139--150},
  year={2025}
}
